\documentclass[letterpaper, 10 pt, conference]{ieeeconf}
\usepackage{graphics} 
\usepackage{epsfig} 
\usepackage{multirow}
\usepackage{rotating}
\usepackage{colortbl}
\usepackage[table, dvipsnames]{xcolor}
\usepackage{makecell}
\usepackage{amssymb}
\usepackage{tabularx,booktabs}
\usepackage{mathtools}
\usepackage{graphicx}
\usepackage{caption}
\usepackage{subcaption}
\usepackage{amsmath} 
\usepackage{bbm}
\usepackage{color,soul}
\usepackage{tikz}
\usepackage{float}
\usetikzlibrary{positioning}
\usepackage{forest,array}
\usetikzlibrary{calc,shapes,shapes.arrows,arrows,trees,shadows,backgrounds,positioning}
\usepackage{multicol}
\usepackage{balance}
\usepackage{xcolor}
\usepackage[framemethod=tikz]{mdframed}
\usepackage{algpseudocode}
\usepackage{algorithm}
\usepackage{bbm}
\usepackage{dsfont}
\usepackage{lipsum}

\forestset{
  nodeStyle/.style={
    before typesetting nodes={
      edge=#1,
      for ancestors={
        edge=#1,
        #1,
      },
      #1,
    }
  },
  my edge label/.style={
    edge label={
      node [midway, fill=white, font=\scriptsize] {#1}
    }
  }
}

\IEEEoverridecommandlockouts                              
\title{\LARGE \bf Prediction of SLAM ATE Using an Ensemble Learning Regression Model and 1-D Global Pooling of Data Characterization}
\author{
	Islam Ali$^{1}$, Bingqing (Selina) Wan$^{2}$, and Hong Zhang$^{3}$

	\thanks{$^{1}$Islam Ali is with the Department of Computing Science, University of Alberta, Edmonton, AB T6G 2R3, Canada {\tt\small iaali@ualberta.ca}}%

	\thanks{$^{2}$Bingqing (Selina) Wan is with the Department of Engineering Science, University of Toronto, Toronto, ON, M5S 2E4 {\tt\small b.wan@mail.utoronto.ca }}%

	\thanks{$^{3}$Hong Zhang is with the Department of Computing Science, University of Alberta, Edmonton, AB T6G 2R3, Canada {\tt\small hzhang@ualberta.ca}}%
}
\begin{document}

\maketitle
\thispagestyle{empty}
\pagestyle{empty}

\begin{abstract}
Robustness and resilience of simultaneous localization and mapping (SLAM) are critical requirements for modern autonomous robotic systems. One of the essential steps to achieve robustness and resilience is the ability of SLAM to have an integrity measure for its localization estimates, and thus, have internal fault tolerance mechanisms to deal with performance degradation. In this work, we introduce a novel method for predicting SLAM localization error based on the characterization of raw sensor inputs. The proposed method relies on using a random forest regression model trained on 1-D global pooled features that are generated from characterized raw sensor data. The model is validated by using it to predict the performance of ORB-SLAM3 on three different datasets running on four different operating modes, resulting in an average prediction accuracy of up to 94.7\%. The paper also studies the impact of 12 different 1-D global pooling functions on regression quality, and the superiority of 1-D global averaging is quantitatively proven. Finally, the paper studies the quality of prediction with limited training data, and proves that we are able to maintain proper prediction quality when only 20 \% of the training examples are used for training, which highlights how the proposed model can optimize the evaluation footprint of SLAM systems.
\end{abstract}
\section{Introduction}
Simultaneous localization and mapping (SLAM) is a fundamental building block that gives modern robotic systems the ability to estimate its location while building a map of the navigated environment \cite{aulinas2008slam}. Over the last few decades, SLAM research has evolved significantly in terms of architecture, accuracy, requirements, and challenges \cite{cadena2016past}. One of the major challenges faced by SLAM is robustness and resilience of the system when deployed in the real world \cite{prorok2021beyond}. Robustness of SLAM is the ability of the system to provide acceptable performance when operating under pre-defined conditions. While resilience is the ability of a system to converge to an acceptable performance when operating outside of the pre-defined conditions, which implicitly highlights the importance of having internal error prediction and tolerance mechanisms in SLAM to allow for this convergence to happen effectively \cite{ali2022iros}. For that reason, researchers have directed their attention towards the introduction of integrity indicators of either some blocks in the SLAM pipeline \cite{carson2022predicting}, or the final SLAM outcome \cite{luperto2021predicting}.

\textit{Absolute Trajectory Error (ATE)} \cite{zhang2018tutorial} is considered the de-facto metric for measuring the accuracy of localization in SLAM and is used by most of state-of-the-art solutions such as ORB-SLAM3 \cite{campos2021orb}, VINS-Mono \cite{qin2018vins}, among many others. Therefore, on-line prediction of SLAM ATE is an integral part of the quest to reach robust and resilient SLAM as it provides SLAM systems with internal indicators of the integrity of their estimates, which can be used to correct estimation errors, govern switching between localization alternatives, and improve robotics safety when deployed. 

In this paper, we propose a novel methodology for predicting the absolute trajectory error (ATE) of a SLAM algorithm using 1-D global pooling of input data characteristics and an ensemble learning-based regression model. This methodology is motivated by the high correlation observed and reported in our previous work \cite{ali2022iros} between the SLAM performance of multiple algorithms on one side, and the characterization metrics measured on different SLAM datasets on the other side. Throughout this work, several design decisions are made such as the selection of the 1-D global pooling function and the generation of sufficient examples for regression training. To ensure proper selection of these choices, a quantitative analysis of the impact of different options on the final accuracy is conducted and presented in this work.

The rest of the paper is organized as follows. Section \ref{sec:related_work} presents a brief review of related work. Then, Section \ref{sec:background} provides a background overview. Next, Section \ref{sec:method} describes our proposed method in detail. After that, results are presented and discussed in Section \ref{sec:results}. Finally, Section \ref{sec:conclusion} presents our conclusions from this study.
\begin{figure*}[!tp]
     \centering
         \includegraphics[width=\textwidth]{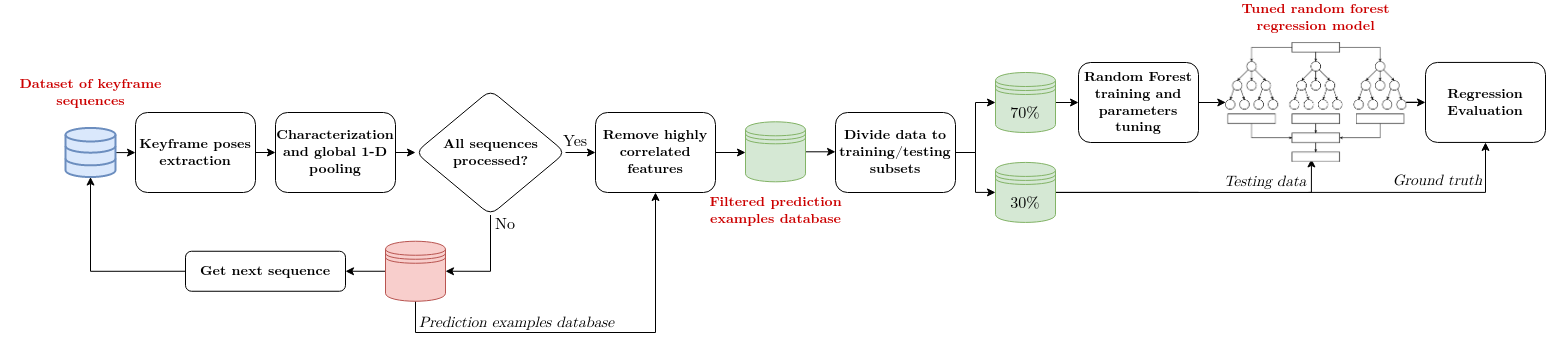}
        \caption{A block diagram of the proposed ATE prediction methodology}
        \label{fig:sys_bd}
\end{figure*}
\section{Related Work}
\label{sec:related_work}
Predicting and estimating system performance is crucial for the safe use of robots and autonomous systems and has been extensively studied in closely related disciplines. For example, navigation systems like INS/GPS use statistical models to estimate errors in sensor measurements and improve localization through Kalman filters \cite{sukkarieh1998achieving} \cite{noureldin2008performance}. Additionally, integrity measures for robot localization have been proposed to minimize deployment risks in real-world scenarios \cite{arana2019recursive}-\cite{hafez2020quantifying}. Despite being essential for robust and resilient perception and navigation, there is limited research on integrity of localization outcomes of SLAM due to unclear design requirements, as reported in \cite{yasuda2020autonomous}. 

The existing literature on integrity measures for SLAM has primarily focused on two areas. The first area is predicting the overall performance of the entire SLAM pipeline. This approach aims to evaluate the integrity of the entire SLAM pipeline as a whole and provide a single measure for SLAM output integrity. For instance, the travelled path of a robot is modeled using Voronoi Graphs to train a model for predicting SLAM performance \cite{luperto2021predicting}. Training data are acquired using the method outlined in \cite{amigoni2018improving} which generates training examples by simulating a SLAM algorithm several times on selected environments. The same training examples generation methodology is used in \cite{piazza2022performance} where a number of univariate and multivariate linear regression models are trained to predict the normalized relative translational error and the absolute trajectory errors. The two approaches proposed are useful in determining the overall performance of SLAM but would not allow for on-line prediction of localization integrity. That is due to their reliance on an overall descriptor of the whole path the robot will traverse rather than incremental raw sensor data (e.g. images). On the other hand, other methods are proposed to identify an upper bound of the localization uncertainty in SLAM providing a guarantee of the system performance when the spatial distribution of features is known \cite{mourikis2006predicting}.

The second area is investigating the integrity of specific components within the SLAM pipeline. This approach aims to evaluate the integrity of individual SLAM components and how the quantification of this integrity measure can be utilized to properly correct potential anomalies in localization estimates. For examples, in \cite{carson2022predicting} a learning-based integrity measure for visual loop closure is proposed to decrease false positives and ultimately improve the overall performance of SLAM localization accuracy through the reduction of loop closure false positives. 

Our approach is unique in both its design and goal. We use a sequence of key-frame measurements such as images and/or inertial measurements as inputs, which are characterized to generate a corresponding characterization matrix of the sequence traversed. Then, we apply a 1-D global pooling function on the rows of the characterization matrix, which results in a 1-D vector descriptor of the sequence. After that, the descriptor is sent to a prediction model to predict the expected ATE at the end of the input sequence. Consequently, this approach allows for on-line monitoring of the system performance, and provides an integrity measure of localization estimates at any time. This is essential for ensuring the robustness and resilience of the SLAM system, particularly in challenging environments or under uncertain conditions.

\section{Background}
\label{sec:background}
To provide a foundation for this study, this section explores two concepts which are: 1-D global pooling and random forests regression. We discuss how these techniques are used and examine their applicability to the proposed work.
\subsection{1-D Global Pooling}
This technique was introduced in \cite{lin2013network} as a solution to the problem of overfitting in neural networks. The technique does this by reducing the spatial dimension of a feature map to a single value using a global pooling function (e.g. average, min, max ...etc.) across all features. Essentially, this reduction replaces a detailed feature map with an abstract, descriptive characteristic of it. Learning those characteristics instead of the examples themselves was proven to enhance the generalization of the learnt model \cite{goodfellow2016deep}. In this work, we utilize this technique and examine the impact of 12 different global pooling functions on prediction quality. Moreover, we examine the impact of concatenating the outcomes of all global pooling functions into a single extended feature vector which provides the learning algorithm with more abstract information about input examples.

\subsection{Random Forest Regression}
Random forest is an ensemble learning technique that relies on the concept of bagging \cite{sutton2005classification} where several decision tree prediction models are trained on independent random sub-samples of the input features in the bootstrapping phases. Bootstrapping is a statistical technique which involves random sub-sampling of the training data pool while allowing replacement \cite{efron1992bootstrap} to generate bootstraps. For each decision tree in the random forest, a bootstrap is selected for training on a random sub-set of available descriptor features. The outcomes of all decision trees are then combined either by averaging (regression) or by majority voting (classification) \cite{breiman2001random}. Due to the independence and low correlation among decision trees, the prediction error is not accumulated or propagated, thus resulting in a lower prediction error.
Random forests provide a way to balance accuracy and generalization, and was proven to be superior to competing methods. For instance, they were proven to outperform neural networks \cite{grinsztajn2022tree} on tabular structured data, and handle overfitting well when compared to boosting algorithms \cite{opitz1999popular}.

\section{Methodology}
\label{sec:method}
This section presents the proposed methodology to predict ATE in SLAM, along with an overview of the design choices made in this work. Figure \ref{fig:sys_bd} illustrates the proposed pipeline and shows how the different system components interact with each other. The figure provides a visual representation of the flow of data and processes in the proposed approach.

Given a dataset $\mathcal{D}$ of $N$ sequences, defined as:
\begin{equation}
\mathcal{D} = \{(\mathbf{x_i}, y_i), i=1,..., N\}
\end{equation}

\noindent where $\mathbf{x_i}$ is a characterization matrix of size $(m \times n)$ corresponds to $m$ characterization metrics applied on an input sequence of $n$ measurements (e.g. images/inertial measurements) and $y_i$ is a scalar corresponding to the ATE of the trajectory.

Each characterization matrix $\mathbf{x_i}$ is transformed to a 1-D vector $\mathbf{z_i}$ of size $(m \times 1)$ by applying a 1-D global pooling function $f(.)$ as such:
\begin{equation}
\mathbf{z_i} = f(\mathbf{x_i})
\end{equation}

Consequently, the transformed dataset $\mathcal{D}'$ is defined as:
\begin{equation}
{\mathcal{D}}' = \{(\mathbf{z_i}, y_i), i=1,..., N\}
\end{equation}

We seek to learn a model to predict SLAM ATE $\hat{y_i}$ given an unseen 1-D vector $\mathbf{z_i}$ from the transformed dataset $\mathcal{D}'$.

\subsection{Data example generation}
\begin{figure}[!tp]
     \centering
         \includegraphics[width=\columnwidth]{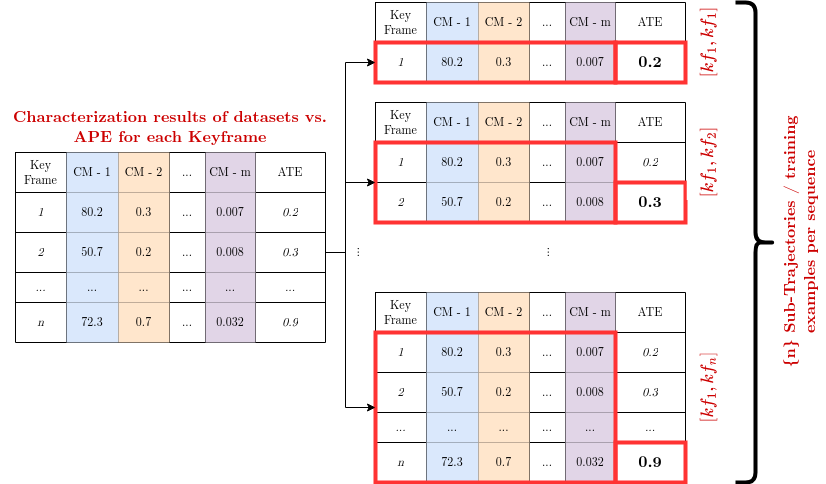}
        \caption{Extraction of training examples based on utilizing sub-trajectory data and corresponding ATE}
        \label{fig:data_creation}
\end{figure}
To generate examples for training the ATE prediction model, we run a SLAM algorithm on all sequences available in several datasets. For each of the run sequences, a number of sub-sequences are calculated that corresponds to the keyframes selected by the SLAM algorithm. Thus, we utilize the concept of sub-trajectories \cite{zhang2018tutorial} in order to expand the number of training examples.
 
Given an input sequence of size $\mathcal{K}$ keyframes, we can extract $\mathcal{K}$ examples, where each example is a sub sequence of keyframes ($kf$) in the inclusive range of $[(kf)_1, (kf)_k]$ where $k=\{1,2 ... \mathcal{K}\}$. The corresponding ATE of each sub trajectory is calculated and is associated with each trajectory to construct a training example. Figure \ref{fig:data_creation} illustrate the process in detail and shows how the training examples extraction and ATE association take place. Sub-trajectories used for model training and testing are generated sequentially from available data sequences in a dataset running in a specific operation mode. The split of the available data for training and testing is done without randomization, meaning that training and testing are conducted on sub-sequences generated from the same dataset but from different sequences.

\subsection{Sequence characterization and 1-D global pooling}
\begin{figure}[!tp]
     \centering
         \includegraphics[width=\columnwidth]{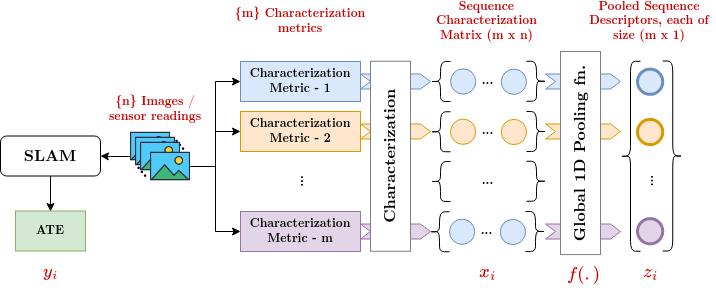}
        \caption{Generation of feature vectors using input sub-sequence characterization and 1-D global pooling}
        \label{fig:data_features}
\end{figure}

Each generated sub-sequence is considered to be an independent sequence of images/sensor readings. We apply the characterization framework introduced in \cite{ali2022iros} which contains an array of characterization metrics (e.g. measuring brightness, contrast ... etc.) that generate a characterization vector for each image/sensor reading in the sequence. As seen in Figure \ref{fig:data_features}, characterization generates a 2D matrix of size $(m \times n)$, where each row represents a characterization metric outcome, and each column represents an input sub-sequence. 
Due to the variability in sequence sizes, the generated 2D matrices are not of the same dimension. Thus, to reduce the dimensionality and provide unified feature vectors for training, we apply a 1-D global pooling function on 2-D matrices to generate 1-D vectors of unified size of $(m \times 1)$. This is achieved by reducing each row in the characterization matrix into a single scalar value using the pooling function. In this work, we utilize one of 12 different pooling functions that include statistical pooling functions (e.g. mean, min, max ... etc.) and diversity pooling functions (e.g. entropy, simpson diversity index and its variants). In order to provide the prediction model with more descriptive features, we also concatenate all 1-D global pooled features into a single feature vector, study its impact on the prediction quality, and compare its performance to using a single 1-D global pooling function. 
\begin{table}[]
\caption{Tuned Hyperparameters in the random forest and their corresponding ranges}
\label{tab:rf_params}
\centering
\begin{tabular}{@{}lc@{}}
\toprule
\multicolumn{1}{c}{\textbf{Hyperparameter}} & \textbf{Range}   \\ \midrule
Number of tree estimators                   & [10, 1000]       \\
Minimum sample required for a split         & \{2,5,10\}         \\
Minimum samples required at a leaf          & \{1,2,4\}          \\
Maximum number of features used for a split & \{None, sqrt, log2\} \\
Maximum depth a tree can grow up to         & [10,100]         \\
Bootstrapping for tree building             & \{True, False\}    \\ \bottomrule
\end{tabular}
\end{table} 
\subsection{Removal of highly correlated features}
The existence of collinearity between independent input features is a potential problem in regression and can lead to numerically unstable results \cite{daoud2017multicollinearity}. To detect highly correlated features, we calculate Pearson correlation coefficient (PMCC) \cite{benesty2009pearson} between each feature and all other available features. After that, highly correlated features are grouped where the PMCC between any two features in a group is greater than a threshold of $95 \%$. Then, only one feature is selected from each group which is then used for the training of the regression model in order to ensure prediction stability of the trained regression model.

\subsection{Random forest regression model}
A random forest regression model is trained and tuned on 70 \% of the data examples available for each test case. After that, the model is tested on the remaining unseen 30 \% in order to determine its performance. We utilize the random forest regression implementation provided in scikit learn library \cite{pedregosa2011scikit} due to its efficiency and ease-of-use. It also exposes a number of hyperparameters that we can tune for optimal performance of the model.

Tuning the random forest hyperparameters is essential to achieve the best prediction performance. For that, we perform a randomized grid search with cross validation on the multi-dimensional space of hyperparameters provided in Table \ref{tab:rf_params}. This method is proven efficient in selecting the best hyperparameters while maintaining reasonable complexity and execution time \cite{probst2019hyperparameters}.

\subsection{Performance Evaluation}
To quantitatively evaluate the regression quality of our method, four different metrics are utilized, which are defined as follows.
\begin{enumerate}
\item Coefficient of determination ($R^2$)
\begin{equation}
R^2 = 1- \frac{\sum_{i=1}^{n} (\hat{y_i} - y_i)^2}{\sum_{i=0}^{n} (y_i - \bar{y_i})^2}
\end{equation}
\item Mean absolute percentage error ($MAPE$)
\begin{equation}
MAPE = \frac{100\%}{n} \sum_{i=1}^{n} \left| \frac{y_i-\hat{y_i}}{y_i} \right|
\end{equation}
\item Mean absolute error ($MAE$)
\begin{equation}
MAE = \frac{1}{n} \sum_{i=1}^{n} \left| y_i-\hat{y_i} \right|
\end{equation}
\item Root mean squared errors ($RMSE$)
\begin{equation}
RMSE = \sqrt{\frac{\sum_{i=1}^{n} (y_i-\hat{y_i})^2}{n}}
\end{equation}
\end{enumerate}
Where $y$ is the ATE ground truth, $\hat{y}$ is the predicted ATE, and $n$ is the number of testing samples.

Those metrics differ in terms of their allowable range, and their indication of the quality of performance. Together, they give a clear indication of the performance and suppress any corner case or anomalies any metric can suffer from.
\section{Experimental Results and Discussion}
\label{sec:results}
In this section, we describe and discuss our experimental setup and associated experimental results. As mentioned in Section \ref{sec:method}, we run ORB-SLAM3 \cite{campos2021orb} on three different datasets and in four different modes of operations, resulting in 10 test cases as illustrated in Table \ref{tab:data_diversity}. For each test case, we examine each of the 12 different 1-D pooling function, train and tune the random forest, and evaluate the model performance. Additionally, we study the impact of reducing the amount of training data on the ATE prediction quality to show how our proposed prediction model can still perform relatively well when limited data is available for training.

The experimental results show that the proposed method is able to predict SLAM ATE with an accuracy up to 94.7\% which is a direct indication of the efficacy of our method, the validity of using the characterization metrics as data descriptors, and the proper choice of the 1-D global pooling function for the SLAM ATE prediction task. 

\subsection{Training data generation}
\begin{table}[!tp]
\centering
\caption{The number of sub-trajectories available from each dataset at different operation mode. }
\label{tab:data_diversity}
\begin{tabular}{@{}lccccc@{}}
\cmidrule[\heavyrulewidth](l){3-6}
                                        & \multicolumn{1}{l}{}    & \multicolumn{4}{c}{\textbf{\# sub-trajectories available}} \\ \toprule
\multicolumn{1}{l|}{\textbf{Dataset}} & \multicolumn{1}{l|}{\textbf{\# Seq}} & \textbf{M} & \textbf{S} & \textbf{M-I} & \textbf{S-I} \\ \toprule
\multicolumn{1}{l|}{\textbf{KITTI}}     & \multicolumn{1}{c|}{22} & 11799        & 23201        & -           & -           \\
\multicolumn{1}{l|}{\textbf{EuroC-MAV}} & \multicolumn{1}{c|}{11} & 3348         & 1956         & 3043        & 1484        \\
\multicolumn{1}{l|}{\textbf{TUM-VI}}    & \multicolumn{1}{c|}{28} & 2049         & 1161         & 4230        & 1924        \\ \bottomrule
\multicolumn{6}{l}{\makecell[l]{* M, S, M-I, and S-I refer to monocular, stereo,\\ monocular-inertial, and stereo-inertial respectively.}}\\

\end{tabular}
\end{table}
To generate examples for training the ATE prediction model, we ran ORB-SLAM3 \cite{campos2021orb} on all sequences available in three different datasets, which are: KITTI \cite{geiger2012we}, EuroC-MAV \cite{burri2016euroc}, and TUM-VI \cite{schubert2018tum}. We apply our proposed data example generation process, which resulted in a great increase in the number of available examples for training, testing, and validation. The method is applied on four different modes of ORB-SLAM3 \cite{campos2021orb} which are monocular, monocular-inertial, stereo, and stereo-inertial, resulting in 10 different test cases. Table \ref{tab:data_diversity} shows the number of training examples generated for each of the test cases.

\subsection{Selection of regression algorithm}
\begin{figure}[!tp]
     \centering
         \includegraphics[width=\columnwidth]{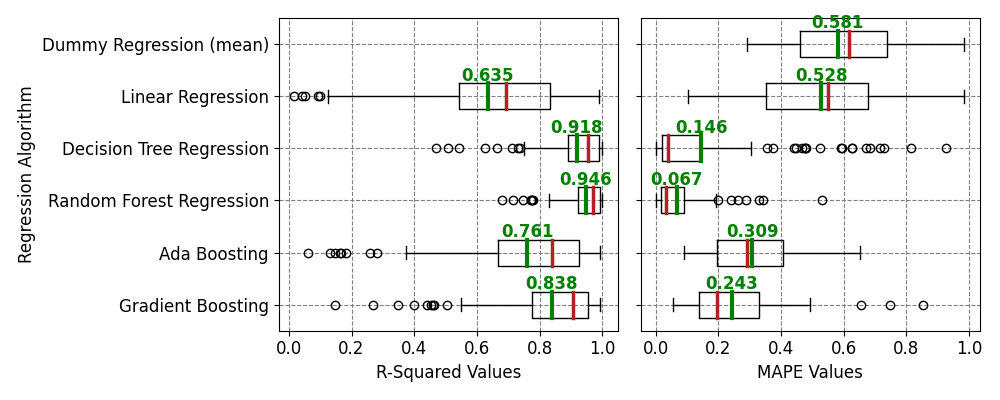}
        \caption{Quantitative comparison of different regression Models for ATE prediction}
        \label{fig:algo_eval_boxplot}
\end{figure}
\begin{figure}[!tp]
     \centering
         \includegraphics[width=\columnwidth]{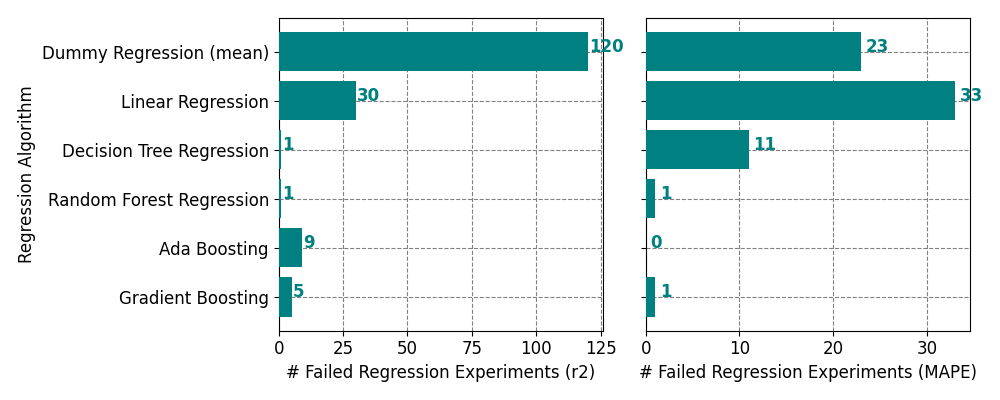}
        \caption{Failure rate statistics of different regression models}
        \label{fig:algo_eval_barplot}
\end{figure}
In order to validate our selection of the regression algorithm, we examined a number of famous regression models with their default hyperparameter values provided in \cite{pedregosa2011scikit}. These algorithms are: dummy regression that takes the average of input features, linear regression, decision tree, random forest, Ada boosting, and gradient boosting. The evaluation is done using $R^2$ and $MAPE$ metrics to allow comparison of different test cases as they provide an absolute measure of performance regardless of the value and range of the predicted variable. 
\begin{figure}[!tp]
     \centering
         \includegraphics[width=\columnwidth]{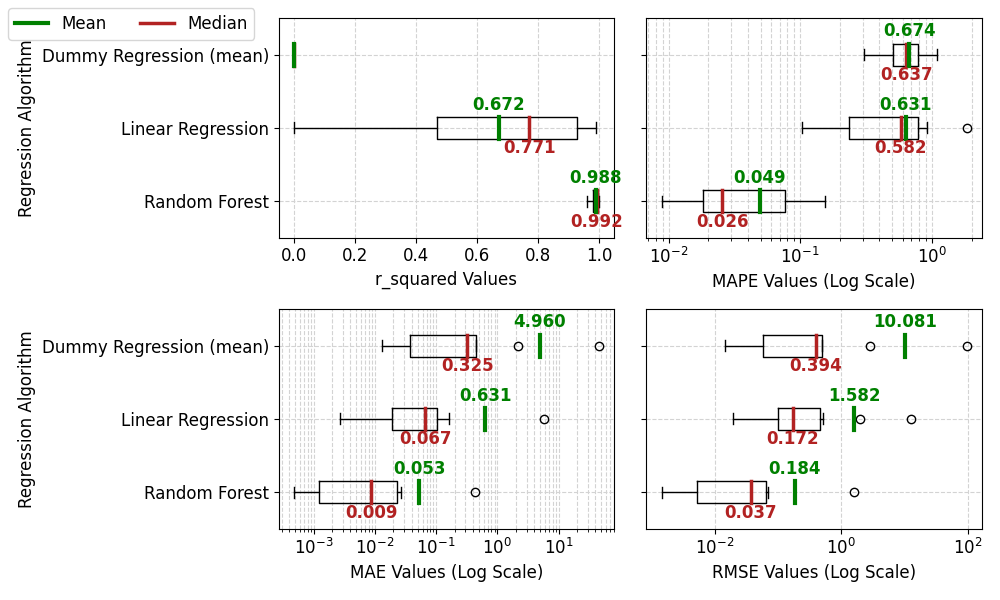}
        \caption{Comparison of random forest regression model performance vs. selected baseline models}
        \label{fig:baseline_compare}
\end{figure}
As shown in Figure \ref{fig:algo_eval_boxplot}, random forests outperform other regression algorithms resulting in the highest $R^2$ value and the lowest $MAPE$ value as well. Additionally, we can clearly observe the overfitting problem of boosting algorithms \cite{opitz1999popular} when we examine Ada boosting and gradient boosting performance compared to random forests. 
\begin{figure*}[!tp]
     \centering
     \begin{subfigure}[b]{0.49\textwidth}
         \centering
         \includegraphics[width=\columnwidth]{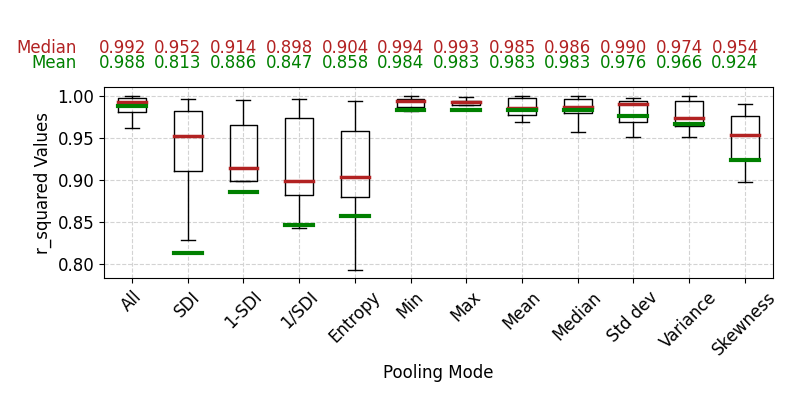}
     \end{subfigure}
     \hfill     
     \begin{subfigure}[b]{0.49\textwidth}
         \centering
         \includegraphics[width=\columnwidth]{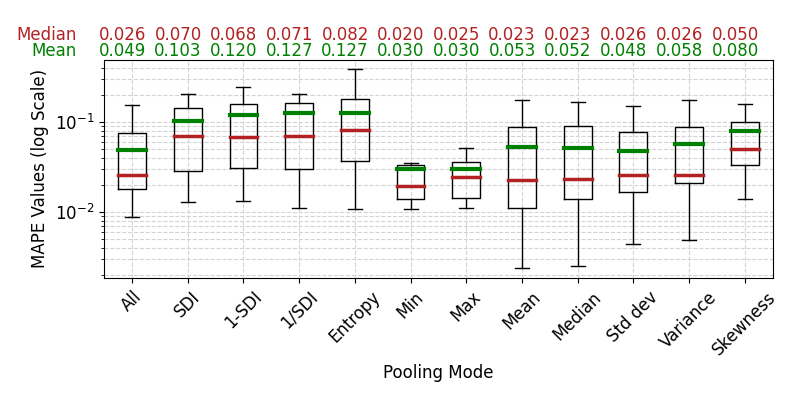}
     \end{subfigure}      
	 \\
     \begin{subfigure}[b]{0.49\textwidth}
         \centering
         \includegraphics[width=\columnwidth]{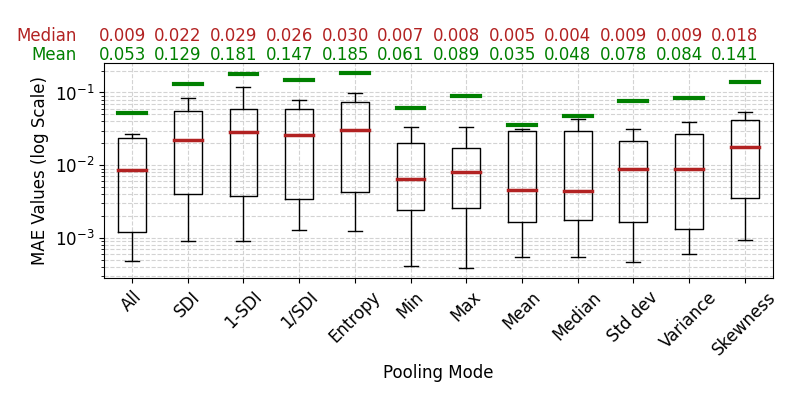}
     \end{subfigure}
     \hfill     
     \begin{subfigure}[b]{0.49\textwidth}
         \centering
         \includegraphics[width=\columnwidth]{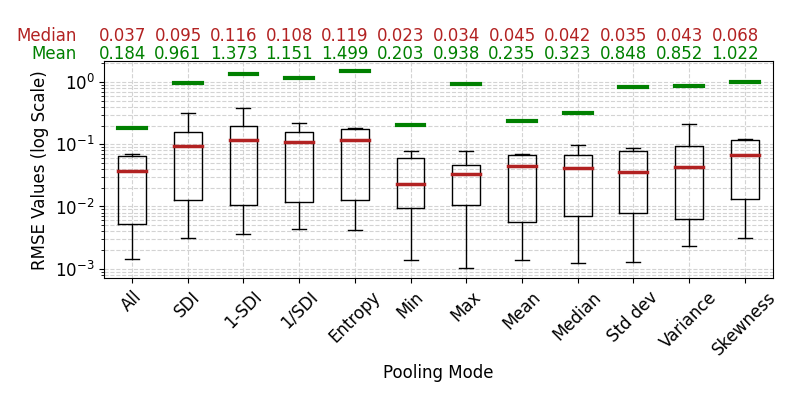}
     \end{subfigure}
        \caption{Comparison of regression quality for different 1-D global pooling functions after training on 70 \% of the data}
        \label{fig:pooling_results}
\end{figure*}
Another experiment included identifying if the algorithm is able to perform regression on a given test case. For that, we utilize the out-of-range concept for both $R^2$ and $MAPE$ as an indication for failed regression as such: $R^4 \in [0,1]$ and $MAPE \in [0,1]$. In Figure \ref{fig:algo_eval_barplot}, we can observe that random forests had the least amount of failed regressions when considering both metrics compared to other algorithms. In fact, we observed that the failure case for random forests happen when testing ORB-SLAM3 in stereo mode on TUM-VI due to the lack of enough samples for training. However, with further tuning, we are able to perform regression on this test cases as well.

\subsection{Performance comparison to baseline}
A standard practice in machine learning is to compare regression results to a baseline model to prove the integrity of the learning process \cite{bhagwat2019applied}. For that reason, we compare the outcomes of the random forest model to two different baselines. The first is a simple dummy regression model that uses the mean of the input features as the regression output. While the second is a linear regression model with default hyperparameter settings. 

As shown in Figure \ref{fig:baseline_compare}, we can observe the superiority of the random forest model compared to the two baseline models in terms of all considered performance metrics, which is a direct indication of the value of using our model for the task in hand. 

\subsection{Impact of limited training data on ATE prediction}
\begin{figure}[!tp]
     \centering
     \begin{subfigure}[b]{0.49\columnwidth}
         \centering
         \includegraphics[width=\columnwidth]{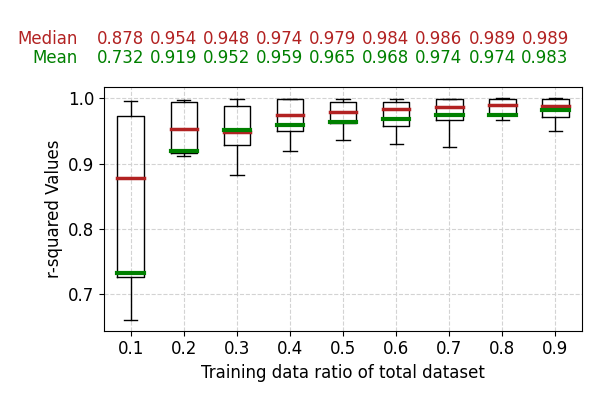}
     \end{subfigure}
     \hfill     
     \begin{subfigure}[b]{0.49\columnwidth}
         \centering
         \includegraphics[width=\columnwidth]{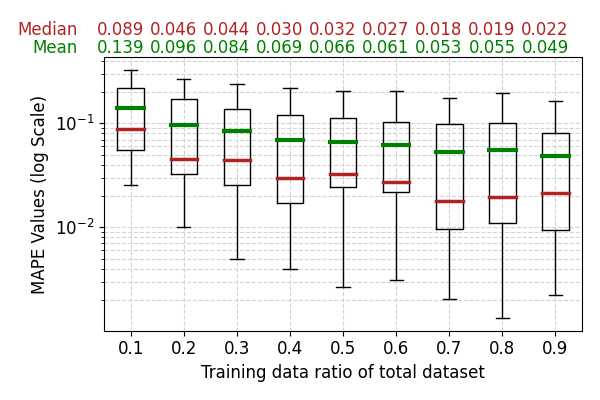}
     \end{subfigure}      
	 \\
     \begin{subfigure}[b]{0.49\columnwidth}
         \centering
         \includegraphics[width=\columnwidth]{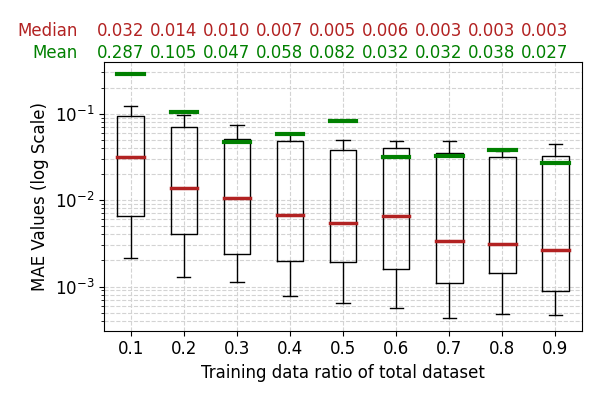}
     \end{subfigure}
     \hfill     
     \begin{subfigure}[b]{0.49\columnwidth}
         \centering
         \includegraphics[width=\columnwidth]{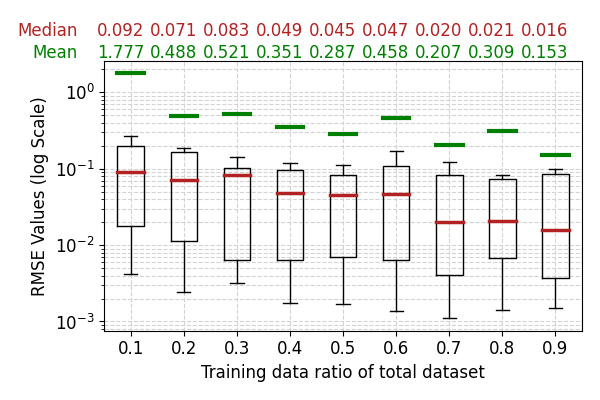}
     \end{subfigure}
        \caption{Effect of reducing training data size on ATE prediction quality using 1-D global average pooling}
        \label{fig:limited_data}
\end{figure}
\begin{table}[!tp]
\caption{Comparison between using the ATE at 20 \% of the trajectory as a predictor for the whole trajectory ATE vs. our proposed random forest method when trained on 20 \% of the available data for all testcases}
\label{tab:vanilla_table}
\centering
\begin{tabular}{@{}rll|ll@{}}
\cmidrule(l){2-5}
\multicolumn{1}{l}{} &
  \multicolumn{2}{c|}{\textbf{Baseline}} &
  \multicolumn{2}{c}{\textbf{Random forest}} \\ \midrule
\textbf{Mode - Dataset} &
  \multicolumn{1}{c}{\textbf{$R^2$}} &
  \multicolumn{1}{c|}{\textbf{$MAPE$}} &
  \multicolumn{1}{c}{\textbf{$R^2$}} &
  \multicolumn{1}{c}{\textbf{$MAPE$}} \\ \midrule
M-EuroC  & 0.1557  & 0.3683 & 0.9973 & 0.0106 \\
M-KITTI  & 0.7621  & 0.3710 & 0.9992 & 0.0079 \\
M-TUMVI  & 0.9036  & 0.2503 & 0.9663 & 0.1246 \\ \midrule
MI-EuroC & -0.0637 & 0.2909 & 0.9888 & 0.0199 \\
MI-TUMVI & 0.8571  & 0.2539 & 0.9835 & 0.1221 \\ \midrule
S-EuroC  & 0.9899  & 0.2699 & 0.9993 & 0.0125 \\
S-KITTI  & 0.9559  & 0.1627 & 0.9999 & 0.0013 \\
S-TUMVI  & 0.6202  & 0.1904 & 0.8458 & 0.0359 \\ \midrule
SI-EuroC & 0.6864  & 0.1954 & 0.9900 & 0.0189 \\
SI-TUMVI & 0.9381  & 0.1681 & 0.9723 & 0.1979 \\ \midrule
\textbf{Mean} &
  \textbf{0.6805} &
  \textbf{0.2521} &
  \textbf{0.9743} &
  \textbf{0.0552} \\ \bottomrule
\end{tabular}
\end{table}
Not surprisingly, reducing the data available for training will reduce the quality of ATE prediction. However, the question is how much reduction should one expect in case of having limited amount of data available for training. In addition to that, we seek to prove that the proposed method can be utilized to reduce evaluation efforts of SLAM by training on a small portion of the dataset and the prediction of the rest of the dataset. \\

In this experiment, we fixed the pooling function to be 1-D global average pooling due to its superior performance compared to other pooling functions. We vary the ratio of training data relative to a testcase from $10 \%$ to $90 \%$ and measure the four regression quality metrics. \\

As shown in Figure \ref{fig:limited_data}, we can observe the normal behaviour of increased prediction quality when more training data is utilized. When we look at the $R^2$ and $MAPE$ metrics, we can observe that we are able to properly predict ATE while training on only $20 \%$ of each test case. In that case, the reduction in $R^2$ is limited to only $6.51 \%$ on average. On the other hand, $MAPE$ also dropped by only $4.7 \%$. Moreover, we can also observe that utilizing less than $20 \%$ of the data for training will produce unreliable results which is indicated by out-of-range values of $R^2$ metric. This experiments directly implies that we are able to reduce the evaluation efforts of SLAM by $80 \%$ while maintaining a decent level of confidence in ATE prediction quality.

Due to the nature of the ATE error and how it evolves over time, we aspire to compare our model when trained on only 20 \% of the data to the ATE observed after traversing 20 \% of the a given data example. This experiments highlights the need for training a prediction model to predict the ATE as the observed ATE during the course of the trajectory is not correlated to that at the end of the trajectory. For that, we compute $R^2$ and $MAPE$ between the ATE at 20 \% of the trajectory and the ATE at the end of the trajectory. Then, we compare the outcomes with that of our prediction model. As shown in Table \ref{tab:vanilla_table}, we can observe that the prediction model outperforms the baseline (ATE at 20 \%) and provides a more accurate outcomes with higher confidence levels. 

\subsection{Evaluation of 1-D global pooling functions}
Another major component of our proposed solution is the 1-D global pooling of training sequences. As described in Section \ref{sec:method}, we examine 12 different pooling functions where 11 of them are either statistical or diversity indicators, while the last one is the concatenation of the outcomes of all of those functions. 

In order to compare the impact of those functions on the regression quality, all test cases are repeated using each pooling function, and performance is evaluated and compared. Figure \ref{fig:pooling_results} shows the comparison results for each metric and highlights the mean and median performance across all test cases for each pooling function. In this figure, each box in the boxplot is constructed from 10 data points where each point represents one of the testcases outlined in Table \ref{tab:data_diversity}. For each testcase, training is done on 70 \% of the data and testing is done on the remaining 30 \%.

It can be observed that the concatenated version of the pooling functions scored a close performance to both the minimum and mean 1-D global pooling functions. However the mean/average 1-D global function provided a balanced and consistent performance when all evaluation metrics are considered. 

\begin{figure*}[!tp]
     \centering
     \begin{subfigure}[b]{0.19\textwidth}
         \centering
         \includegraphics[width=\columnwidth]{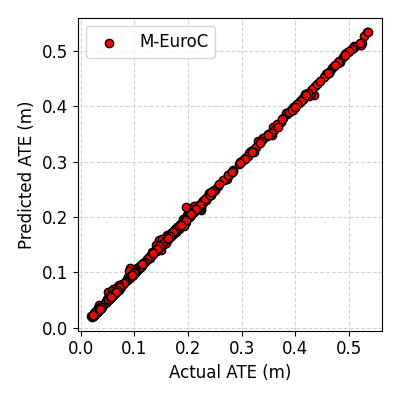}
     \end{subfigure}
     \hfill     
     \begin{subfigure}[b]{0.19\textwidth}
         \centering
         \includegraphics[width=\columnwidth]{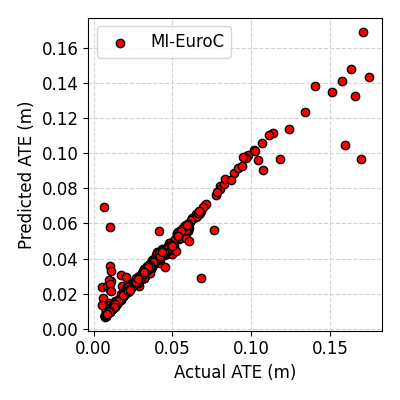}
     \end{subfigure}   
     \hfill     
     \begin{subfigure}[b]{0.19\textwidth}
         \centering
         \includegraphics[width=\columnwidth]{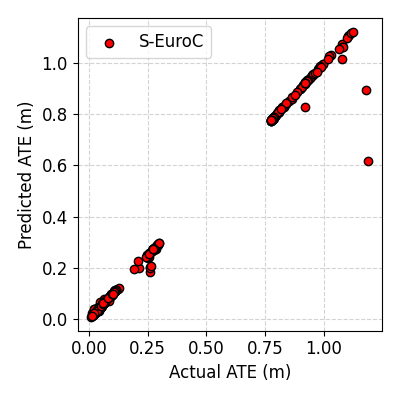}
     \end{subfigure}  
     \hfill     
     \begin{subfigure}[b]{0.19\textwidth}
         \centering
         \includegraphics[width=\columnwidth]{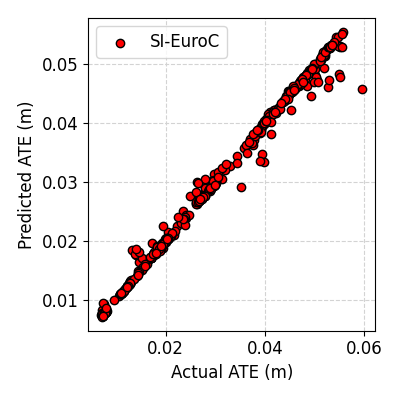}
     \end{subfigure}  
     \hfill     
     \begin{subfigure}[b]{0.19\textwidth}
         \centering
         \includegraphics[width=\columnwidth]{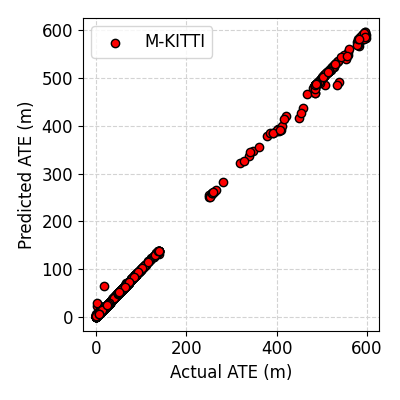}
     \end{subfigure}    
	 \\
     \begin{subfigure}[b]{0.19\textwidth}
         \centering
         \includegraphics[width=\columnwidth]{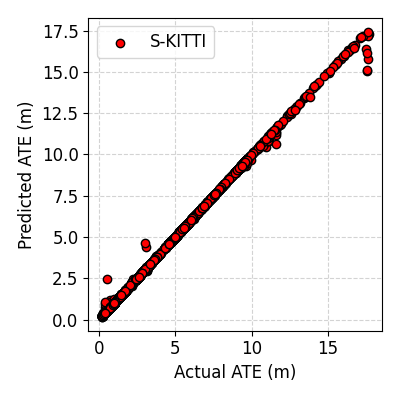}
     \end{subfigure}
     \hfill     
     \begin{subfigure}[b]{0.19\textwidth}
         \centering
         \includegraphics[width=\columnwidth]{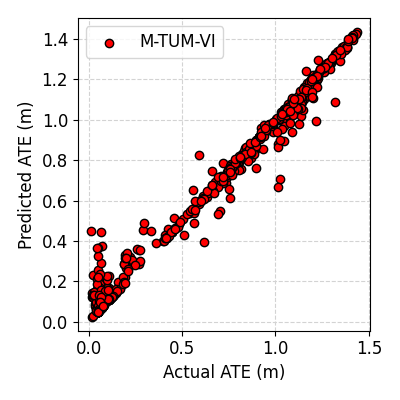}
     \end{subfigure}   
     \hfill     
     \begin{subfigure}[b]{0.19\textwidth}
         \centering
         \includegraphics[width=\columnwidth]{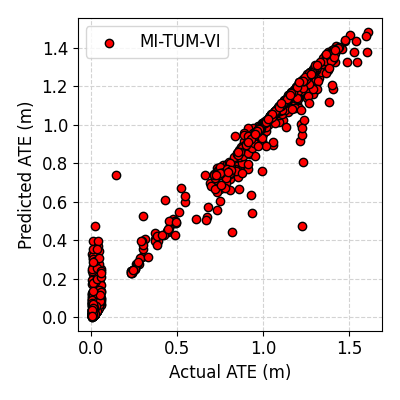}
     \end{subfigure}  
     \hfill     
     \begin{subfigure}[b]{0.19\textwidth}
         \centering
         \includegraphics[width=\columnwidth]{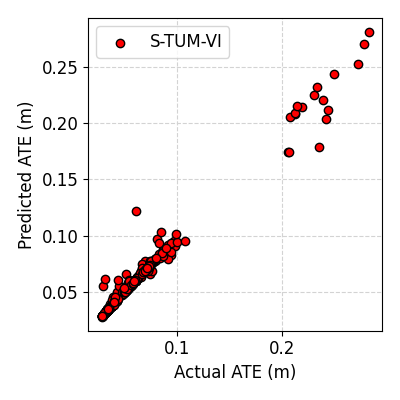}
     \end{subfigure}  
     \hfill     
     \begin{subfigure}[b]{0.19\textwidth}
         \centering
         \includegraphics[width=\columnwidth]{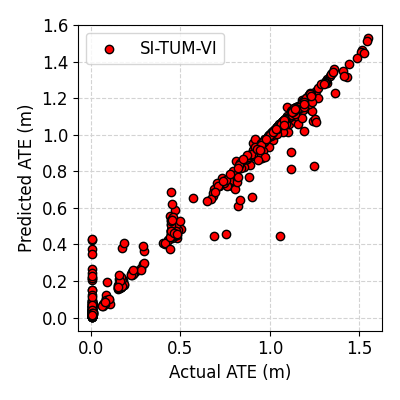}
     \end{subfigure}   
        \caption{Actual vs. predicated ATE for all evaluated testcases using the 1-D global average pooling and random forests after training on 70 \% of the data}
        \label{fig:output_summary}
\end{figure*}

\begin{figure*}[!tp]
     \centering
     \begin{subfigure}[b]{0.19\textwidth}
         \centering
         \includegraphics[width=\columnwidth]{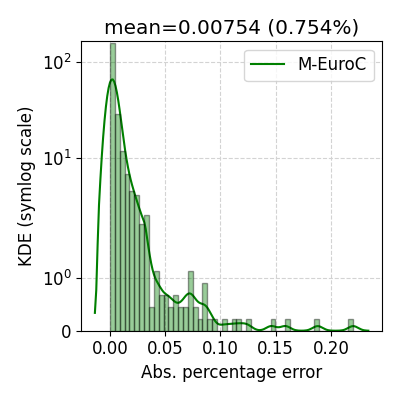}
     \end{subfigure}
     \hfill     
     \begin{subfigure}[b]{0.19\textwidth}
         \centering
         \includegraphics[width=\columnwidth]{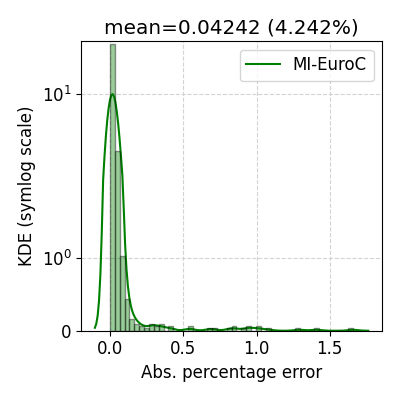}
     \end{subfigure}   
     \hfill     
     \begin{subfigure}[b]{0.19\textwidth}
         \centering
         \includegraphics[width=\columnwidth]{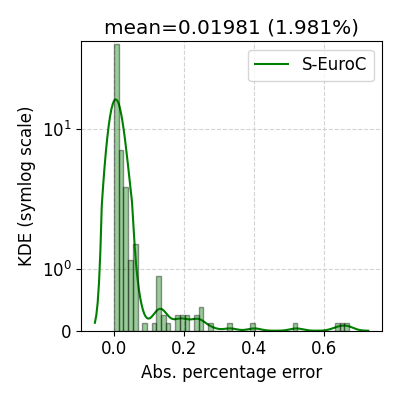}
     \end{subfigure}  
     \hfill     
     \begin{subfigure}[b]{0.19\textwidth}
         \centering
         \includegraphics[width=\columnwidth]{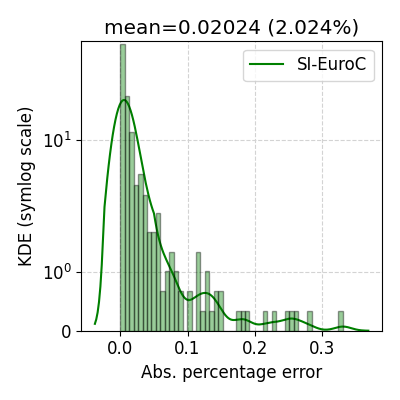}
     \end{subfigure}  
     \hfill     
     \begin{subfigure}[b]{0.19\textwidth}
         \centering
         \includegraphics[width=\columnwidth]{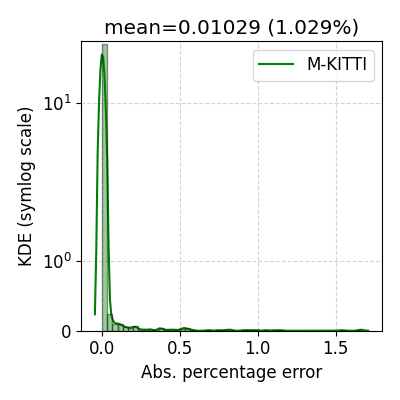}
     \end{subfigure}    
	 \\
     \begin{subfigure}[b]{0.19\textwidth}
         \centering
         \includegraphics[width=\columnwidth]{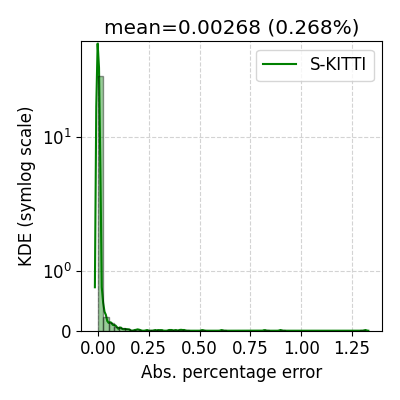}
     \end{subfigure}
     \hfill     
     \begin{subfigure}[b]{0.19\textwidth}
         \centering
         \includegraphics[width=\columnwidth]{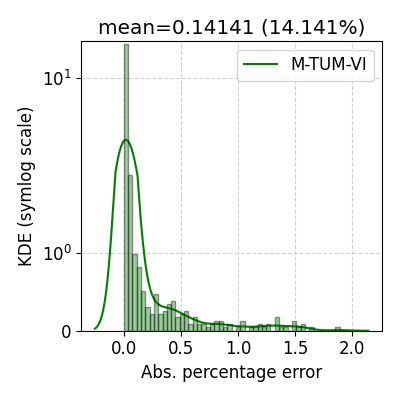}
     \end{subfigure}   
     \hfill     
     \begin{subfigure}[b]{0.19\textwidth}
         \centering
         \includegraphics[width=\columnwidth]{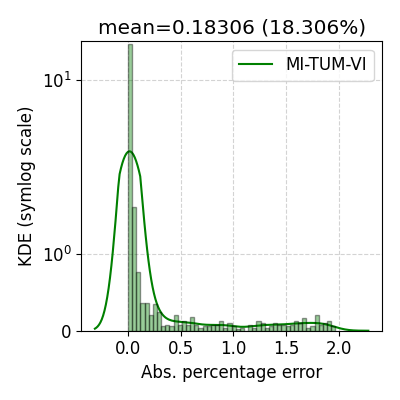}
     \end{subfigure}  
     \hfill     
     \begin{subfigure}[b]{0.19\textwidth}
         \centering
         \includegraphics[width=\columnwidth]{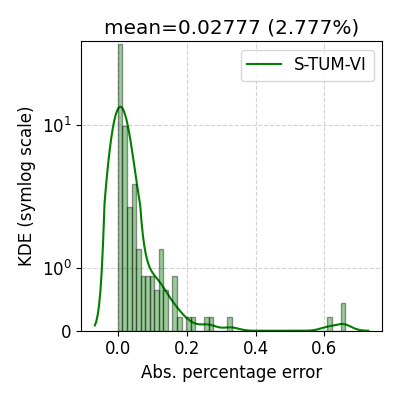}
     \end{subfigure}  
     \hfill     
     \begin{subfigure}[b]{0.19\textwidth}
         \centering
         \includegraphics[width=\columnwidth]{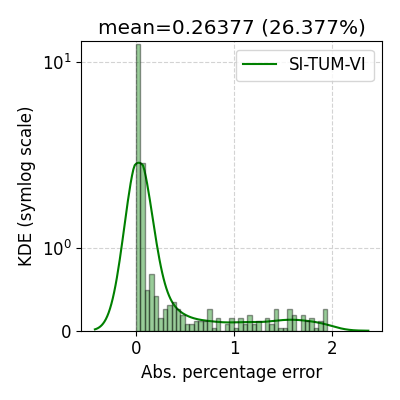}
     \end{subfigure}   
        \caption{Absolute error percentage of all examined testcases using 1-D global average pooling and random forests after training on 70 \% of the data}
        \label{fig:smape_summary}
\end{figure*}

Thus, the comparison between the actual ATE and the predicted ATE using 1-D global average pooling and random forests is presented in Figure \ref{fig:output_summary} for the 10 testcases we examined. Moreover, we present the kernel distribution estimate (KDE) of the absolute error percentage of all testing example in each of the 10 testcases in Figure \ref{fig:smape_summary}. One can observe that we are able to predict the ATE value within an average error of $6\%$ to $7\%$ of the actual ATE with peak performance of an average error that is less than $5 \%$ of the actual ATE value in 7 out of 10 testcases examined. 


\section{Conclusions}
\label{sec:conclusion}
In this paper, the problem of performance prediction in SLAM is addressed as a fundamental requirement for robustness and resilience in SLAM. The study starts by giving a brief review of the literature related to this topic, and provides a basis for the proposed algorithm. After that, we introduce our methodology for predicting SLAM ATE using an ensemble learning technique and 1-D global pooling of input data characterization results. Our methodology is first compared to a multitude of regression models to validate our selection of random forests as our regression model. Then, the methodology is tested on 10 different test cases and using 12 different 1-D global pooling functions, resulting in 120 different experiments. The experimental results showed a superiority in using random forest compared to our selected baseline and provided an evidence for the ability to predict ORB-SLAM3 ATE using characterized and pooled features with accuracy that can reach 94.7\% on average for test cases examined. The paper also evaluated the impact of the selection of a certain pooling function on the regression quality of ATE. It is shown that very limited gain is observed by concatenating multiple pooling functions. It also provided evidence on the suitability of using 1-D global averaging for achieving a good balance on all performance metrics measured. Additionally, the paper studied the impact of reducing the amount of training data on ATE prediction quality, and it is shown that we are able to use only $20 \%$ and maintain an average absolute percentage error of $90.4\%$ with a degradation of only $4.7\%$ compared to training on $90\%$ of available data. Finally, the study highlights the suitability of the used characterization results to be treated as a data sequence descriptor. 
\bibliographystyle{IEEEtran}
\balance
\bibliography{refs}

\end{document}